# Classifying Organizations for Food System Ontologies using Natural Language Processing

Tianyu JIANG [a], Sonia VINOGRADOVA [b], Nathan STRINGHAM [a],
E. Louise EARL [b], Allan D. HOLLANDER [c], Patrick R. HUBER [c], Ellen RILOFF [a,1],
R. Sandra SCHILLO [b], Giorgio A. UBBIALI [d], Matthew LANGE [e],

[a] *University of Utah*
[b] *University of Ottawa*
[c] *University of California, Davis*
[d] *University of Milan*
[e] *IC-FOODS*

ORCiD ID: E. Louise Earl https://orcid.org/0000-0002-5646-0132, R. Sandra Schillo https://orcid.org/0000-0002-3468-0206, Giorgio A. Ubbiali https://orcid.org/0000-0001-7872-1770 , Matthew Lange https://orcid.org/0000-0002-6148-7962

**Abstract.** Our research explores the use of natural language processing (NLP) methods to automatically classify entities for the purpose of knowledge graph population and integration with food system ontologies. We have created NLP models that can automatically classify organizations with respect to categories associated with environmental issues as well as Standard Industrial Classification (SIC) codes, which are used by the U.S. government to characterize business activities. As input, the NLP models are provided with text snippets retrieved by Google's search engine for each organization, which serves as a textual description of the organization that is used for learning. Our experimental results show that NLP models can achieve reasonably good performance for these two classification tasks, and they rely on a general framework that could be applied to many other classification problems as well. We believe that NLP models represent a promising approach for automatically harvesting information to populate knowledge graphs and aligning the information with existing ontologies through shared categories and concepts.

## 1. Introduction

Food systems include activities and relationships that are involved with the production, transport, and consumption of food, and thus are linked to a wide variety of natural and human systems (Tomich et al. [13]). Their extensive nature can have dramatic impacts on these natural and human systems (IPBES [9], UNEP [15]). Conversely, food systems are vulnerable to changes in these other systems. Food activities also show a pivotal role in fostering human social and health-related issues. For instance, inequalities and injustices exist across the whole global food supply chain, as well as within related sectors (D'Odorico et al. [5], Nicolétis and Termine [10]). To date, hunger, malnutrition, and micronutrient deficiencies, just to cite a few, still largely threaten humanity worldwide (Fanzo et al. [6], Fanzo and Davis [7], UNICEF [14]).

---
[1] Corresponding Author: Ellen Riloff (riloff@cs.utah.edu).

As our world becomes more interconnected and interdependent, ontologies become useful ways of categorizing and relating information together. This is especially applicable to food systems, where data access, interoperability, and reusability are essential to deal with the interrelated issues arising from food systems activities. An ontology is a "formal theory" which provides a "commonly accepted" dictionary of terms, supported by a "canonical syntax" and a set of axioms, for a knowledge domain of interest (Smith [12]). In doing so, an ontology offers a common semantic framework for that domain of knowledge (Smith [12], Hollander et al. [1]). Thus, an ontology fosters data access, interoperability, and reusability across several disparate resources, which employ that ontology as a common reference semantic standard (Hollander et al. [1]). Nowadays, several ontologies focusing on food, health, and related aspects have been developed (Boulos et al. [2], Dooley et al. [3]). Here, we just cite a few noteworthy ones as examples: 1) AGROVOC[2] is a three-decade-long well-established "multilingual thesaurus", addressing food and related domains of interest (Boulos et al. [2]). It belongs to the United Nations Food and Agriculture Organisation (FAO). 2) The Agronomy Ontology AgrO[3] was developed within the frame of the CGIAR Platform for Big Data in Agriculture, and which proposes a vocabulary of terms, covering "agronomic management practices, implements, and variables used during agronomic experiments" (Dooley et al. [3]). 3) The Compositional Dietary Nutrition Ontology CDNO[4] presents terms related to "nutritional attributes from crops, livestock, and fisheries that contribute to human diet and which are referenced in precision food commodity laboratory analytics" (Dooley et al. [3]). 4) The Food Ontology FoodOn[5] provides a lexicon about "basic raw food source ingredients, process terms for packaging, cooking, and preservation, and an upper-level variety of product type schemes under which food products can be categorized". FoodOn stands out as a fundamental ontology in addressing food-related aspects (Dooley et al. [4]). Further, FoodOn aims to state the lingua franca within the domain of food, for sharing and reusing food-related information both by humans and machines (Dooley et al. [4]). Despite the availability of food-health ontologies, the envisaged data access, interoperability and reusability across food industries and other food-related sectors, as we claimed above, appears to have not been reached yet (Tomich et al. [13]).

In this paper, we present new research that uses artificial intelligence (AI) technology to automatically categorize organizations, ultimately for the purpose of linking them into food system ontologies. Our eventual goal is to automatically populate large knowledge graphs of information related to agriculture and food systems, where the concepts in the graph are aligned with well-established ontologies to ensure that the knowledge will be represented consistently and aligned with other resources and systems that rely on the same ontological frameworks. Populating knowledge graphs by hand is time-consuming and expensive, so AI technology offers the opportunity to automatically harvest information much more quickly and efficiently.

---

[2]`https://www.fao.org/agrovoc/`
[3]`https://bigdata.cgiar.org/resources/agronomy-ontology/`
[4]`https://cdno.info/`
[5]`https://github.com/FoodOntology/foodon/`

Specifically, we focus on two classification tasks related to organizations. We aim to categorize organizations based on 1) a set of environmental issues that are relevant to environmental planning and food systems, and 2) standard industrial classification (SIC) codes[6], which the U.S. government assigns to businesses to categorize the nature of their activities. SIC codes are analogous to North American Industry Classification System (NAICS) codes, which are used across North America. We have designed natural language processing (NLP) models that read text associated with an organization to automatically assign the organization to categories for these two tasks. In the following sections, we describe the classification tasks in more detail, explain how we collect relevant texts for the NLP models to use, present the NLP technology underlying the classification models, and show experimental results for the two classification tasks.

## 2. Classification Tasks & Datasets

### 2.1. Environmental Issues Classification Task and Dataset

There is a large amount of information currently available concerning the state of the environment. Around the world, many organizations are collecting and analyzing data. However, there remains a major gap in our ability to connect these data sources and make them "smart". They typically use different formats and vocabularies, rendering them unable to be used together. The conservation community lacks an informatics backbone to begin linking people and data in ways that enhance our capacity for informed decision making in our effort to conserve and enhance the Earth's ecosystems. This need is especially crucial as we enter an unprecedented era of rapid environmental change.

One key question in environmental planning, food systems, and many other contexts is "Who is doing what where?". "Who" can be people or organizations, "what" may be projects or other activities, and "where" could refer to many kinds of geographies. To help provide machine-readable answers to this question, Hollander et al. [1] developed an ontology called "PPOD" (People, Projects, Organizations, and Datasets).[7] This ontology formally describes the characteristics of and relationships between these classes of information.

Members of our team have instantiated the PPOD ontology with information concerning the conservation of working landscapes in California. This knowledge graph (KG) contains over 2,000 organizations which were identified and collected in an ad hoc manner through an array of online searches using terms such as "conservation", "biodiversity", "grazing", and "water supply". Each organization is associated with multiple attributes that describe its structure and mission, such as *hasOrgType*, *hasOrgActivity*, and *issues*. The "*issues*" attribute describes potential environmental issues associated with the organization. PPOD has pre-defined 44 high-level environmental issues called "integrated issues", and 325 more fine-grained environmental issues called "component issues". The ontology provides a detailed textual description for each issue label and the relation between integrated issues and component issues. For example, component issues "Air Pollution", "Air Quality", "Greenhouse Gas Mitigation" and "Greenhouse Gas Emissions" are children of the integrated issue "Air & Climate", which is described as "GHG emissions, ozone layer depletion, air quality, climate change influenced and extreme weather events, shifts in growing zones for key crops due to climate change." Expert opinion was used to associate each organization with one or more issues.

---

[6]`https://www.osha.gov/data/sic-manual`
[7]`https://github.com/PPODschema`

| Category | # of Organizations | % Percentage | Example Organization |
| --- | --- | --- | --- |
| Water | 845 | 39.0 | American Rivers |
| Physical Infrastructure | 702 | 32.4 | Rebuild NorthBay Foundation |
| Wastes & Pollution | 578 | 26.7 | Heal the Ocean |
| Biodiversity | 556 | 25.7 | Feather River Land Trust |
| Land & Soil | 535 | 24.7 | Agricultural Research Service |
| Food Production | 393 | 18.2 | American Grassfed Association |
| Institutions | 345 | 15.9 | Southern California Edison |
| Governance | 274 | 12.7 | Merced County |
| Protected Areas | 253 | 11.7 | American Forest Resource Council |
| Sociocultural Systems | 231 | 10.7 | Enterprise Rancheria |
| Public Health | 179 | 8.3 | California Department of Public Health |
| Disasters | 162 | 7.5 | Tahoe Fire & Fuels Team |
| Common Pool Resources | 128 | 5.9 | Sustainable Conservation |
| Air & Climate | 126 | 5.8 | Irvine Global Warming Group |
| Technology | 122 | 5.6 | CDFW Data and Technology Division |

**Table 1.** Examples of environmental issues and how many organizations each issue is associated with. The last column shows an example organization labeled with each issue category.

We define a task based on the PPOD ontology called **environmental issues classification**. Given an organization, the task requires that the NLP model assign one or multiple **environmental issue category labels** to the organization that describe the organization's activities and/or mission. The NLP model needs to be trained with a reasonable number of examples for each category, so we started with the most common categories in the existing PPOD data. We mapped all component issues to their parent integrated issues (could be more than one), then sorted the issues based on the number of organizations associated with each issue. Finally, we selected the 15 most common integrated issues to be our set of category labels. The resulting dataset contains 1,870 organizations that are associated with 15 environmental issue categories: Air & Climate, Biodiversity, Common Pool Resources, Disasters, Food Production, Governance, Institutions, Land & Soil, Physical Infrastructure, Protected Areas, Public Health, Sociocultural Systems, Technology, Wastes & Pollution, and Water. Table 1 shows the categories, the number and percentage of organizations that each category is associated with, and an example organization for each category.

*2.2. Standard Industrial Classification (SIC) Task and Dataset*

Standard Industrial Classification (SIC) codes were created by the U.S. government to categorize businesses according to the industry that they serve and operate in. We believe that incorporating industry classification such as SIC codes, which remain in use although replaced by the North American Industry Classification System in 1997, into food system ontologies is valuable for understanding the nature of an organization's activi-ties and its economic and logistical relationships with other organizations (e.g., supply chain relationships). Although the SIC codes for many organizations can be looked up in government or business databases, having an AI model that can automatically assign SIC codes to an organization could be used to 1) categorize newly formed organizations quickly, without waiting for official databases to be updated, 2) categorize organizations outside of the U.S. with respect to these standardized industry codes, and 3) maintain the currency of a knowledge graph by automatically reclassifying organizations on a regular basis (say, annually) to reflect changes that an organization has made in its activities (e.g., expansion of business activities, or retraction). Toward this end, we define a second task called **SIC code classification**, which requires the NLP model to assign one or multiple **SIC code category labels** to an organization.

The SIC codes are 4 digits long and are hierarchical. These digits represent the Division, Major Group, Industry Group, and Industry of an entity, respectively. For example, a company with code 0116 would belong to the "Soybeans" industry within the "Cash Grains" industry group and the "Agricultural Production Crops" major group. This system allows us to study companies at different levels of granularity by simply grouping companies according to the first 1,2,3 or 4 digits of their SIC codes.

To train a NLP model for this task, we need examples of organizations and their associated SIC codes. Conveniently, the Securities and Exchange Commission (SEC) maintains a publicly accessible database of U.S. companies called EDGAR. This database contains SIC codes for companies as well as their SEC filing reports. Of particular interest are the 10-K and 20-F forms, which provide a company's annual report. These forms detail a range of information related to the company's operations in the past year, including financial, legal, risk factors, and other information that allows stakeholders to assess the state of the business. So we downloaded these reports as well, as a source of textual information that the NLP model could potentially use. Specifically, we collected the natural language text from the "Item 1: Business" section of a company's most recent **10-K filing**, as well as the company's official SIC code. In the EDGAR database there is only one SIC code per company, despite the fact that in principle a company could have multiple codes associated with it.

To collect company information, we started with a list of all 816,115 companies in the database and their Central Index Key (CIK) number, which is provided by the SEC.[8] Then we queried each CIK number in EDGAR to collect the *name*, *SIC*, *SIC description* and their *10-k forms*. We focused our research on the 36,715 organizations that had all of these fields. We observed that the distribution of the data is highly skewed, with many SIC codes containing very few instances. This may be partly due to the fact that organizations are forced to pick a SIC/NAICS code when incorporating their organization, but business models change, and they may be involved in multiple lines of business that can be reflected by multiple SIC/NAICS codes.

For our experiments, we created a balanced subset of the SEC data so that we have the same amount of information for each category and can fairly compare the performance of our NLP models across categories. We decided to focus on just the first 2 digits of each SIC code as the category labels, which helps to minimize data sparsity (because many of the longer 3-digit and 4-digit codes have relatively few organizations associated with them) and provides a useful high-level view of each company's general type of busi-ness operations. To create the dataset for our experiments, we selected all of the 2-digit SIC codes that have at least 200 associated companies, which resulted in the set of 27 SIC codes shown in Table 2. Finally, we randomly sampled 200 companies for each of these codes, which created a balanced dataset containing 5,400 organizations with their 2-digit SIC codes.

---

[8]https://www.sec.gov/Archives/edgar/cik-lookup-data.txt

| SIC | Description | SIC | Description |
|---|---|---|---|
| 10 | Metal Mining | 58 | Eating and Drinking Places |
| 13 | Oil and Gas Extraction | 59 | Miscellaneous Retail |
| 20 | Food and Kindred Products | 60 | Depository Institutions |
| 27 | Printing, Publishing and Allied Industries | 61 | Nondepository Credit Institutions |
| 28 | Chemicals and Allied Products | 62 | Security |
| 34 | Fabricated Metal Products | 63 | Insurance Carriers |
| 35 | Industrial and Commercial Machinery ... | 65 | Real Estate |
| 36 | Electronic | 67 | Holding and Other Investment Offices |
| 37 | Transportation Equipment | 70 | Hotels, Rooming Houses, Camps ... |
| 38 | Measuring, Photographic, Medical, | 73 | Business Services |
| 48 | Communications | 79 | Amusement and Recreation Services |
| 49 | Electric, Gas and Sanitary Services | 80 | Health Services |
| 50 | Wholesale Trade - Durable Goods | 87 | Engineering, Accounting, Research ... |
| 51 | Wholesale Trade - Nondurable Goods | | |

**Table 2.** Descriptions for the most common SIC codes.

*2.3. Organization Information Collection via Google Search*

Our goal is to create a classification model that is given a text about an organization as input and produces category labels for that organization as output. So a key question is: where can we find text that describes an organization? We initially considered using the official website for an organization as the input text because websites often contain information about an organization's initiatives, policies and practices. However, 1) some organizations do not have an official website, and 2) many websites do not allow automated crawling, and in that case we cannot use a web scraper to automatically extract the text from the website. Of the 2,165 organizations in the environmental issues dataset, we found that only about half of the organizations' websites could be crawled.

As an alternative, we decided to use the Google Search Engine to retrieve textual information about organizations. For each organization, we give the organization's name as a search query to the Google Search API and extract the first 10 returned results. The search results from Google provide several types of information, such as organic results, knowledge graph, local results, related questions, etc. Organic results are the algorithmically calculated query results (as opposed to advertisements) that most users typically read, which includes the title, link, and a text "snippet" for each retrieved web page. We use the text snippet, which is the small block of text that appears underneath the link to a website. It is usually around 100-200 characters in length and typically provides users with a general description of the content of the website. For each organization, we then concatenate the text snippets from the top 10 retrieved websites into a single "pseudo-document" (**PseudoDoc**), which serves as the text data that we use for the organization. Figure 1 shows the pipeline for creating the PseudoDoc for an organization using the Google Search API.

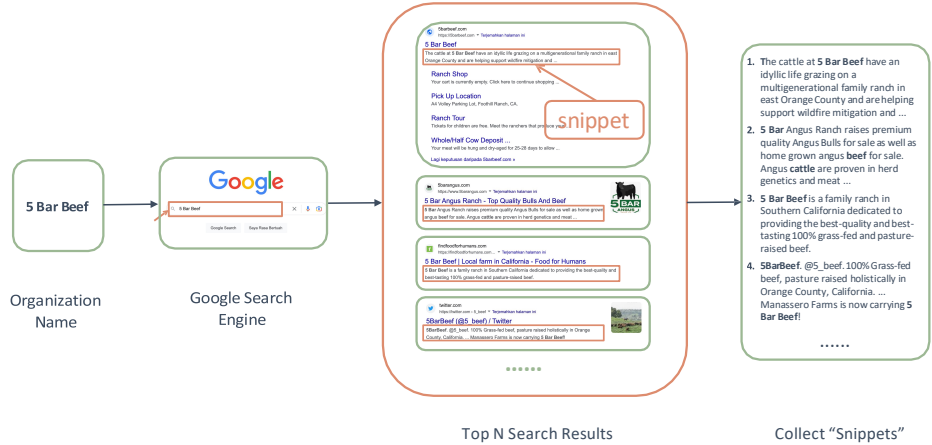

**Figure 1.** Download Google snippets for each organization as its textual representation.

As we will show in Section 4, using these text snippets from retrieved web pages produced reasonably good classification models. We observed two things that explain why Google Search produced useful textual information about organizations. First, when an organization did have an official website, Google typically found it and ranked it as the #1 or #2 top hit. So in many cases, the text snippets returned for an organization include text from the organization's own website. Second, the other websites retrieved by Google for an organization usually either (a) discussed the organization, or (b) discussed similar organizations (i.e., organizations with similar names). Consequently, the text snippets from those websites often contained relevant information that could be useful for inferring the organization's activities and mission. By using text snippets from 10 retrieved websites, each pseudo-document contained information about the organization (or similar organizations) that originated from multiple sources, which collectively painted a good picture of the nature of the organization.

## 3. Methods

### 3.1. Background: Pre-trained Language Models

Pretrained language models, such as BERT [8] and GPT-2 [11], have achieved great success in the field of Natural Language Processing because of their ability to absorb a lot of information about language from massive amounts of text, without any human supervision. These large language models are neural network ("deep learning") architectures that can be additionally trained for a specific application task using a method called **fine-tuning**, where the model is provided with human-labeled data for the application task. During fine-tuning, the model combines the general knowledge about language that it previously absorbed during pre-training with the new information in the task-specific data. Fine-tuned models can perform very well for many application tasks, even when provided with only a small amount of task-specific data.

For this work, we use a well-known pre-trained language model called BERT [8]. BERT has been pre-trained on 3.3 billion English words from Wikipedia and the Google

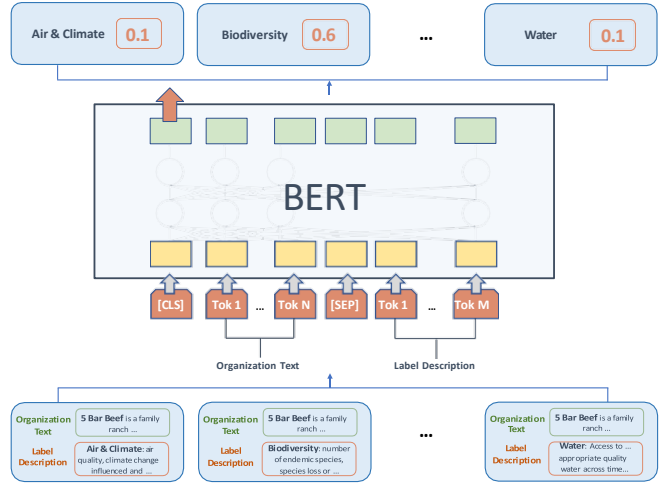

**Figure 2.** OrgModel-2 model architecture and an illustration example for organization *5 Bar Beef*.

Books text collection. We use the base variant of BERT which has 12 layers (transformer blocks), 768 hidden units (hidden size), and 12 self-attention heads. The BERT-base model consists of approximately 110 million parameters (learned weights), offering a good balance between computational efficiency and performance across a wide range of natural language processing tasks.

*3.2. Classification Models*

We created two different designs for our classification models — one basic fine-tuning design and one slightly more complex design. The first model, OrgModel-1, takes the text associated with an organization as input and fine-tunes the BERT language model with the "gold" training data (labeled examples) for the task. Specifically, we follow the common practice of using the embedding vector of the [CLS] token for the classification task and stacking a linear classification layer on top of BERT's last layer, which produces an *n*-dimensional output vector, where *n* is the number of categories for the task.

For the SIC code classification task, we use cross-entropy loss for training. The environmental issues classification task is slightly different because it is a multi-label problem (an organization can be associated with more than one issue), so we further apply a sigmoid function to transform each dimension value to a number between 0 and 1. If the number is $\geq 0.5$, the system predicts *Yes* (meaning the organization belongs to this environmental issue category), otherwise *No*. We use binary cross-entropy loss during training.

The second model, OrgModel-2, takes advantage of an additional source of information: the model is provided with expert-written descriptions for each category as input, along with the text associated with an organization. This provides the model with a definition for each category so that the model can potentially produce a richer semantic representation of the categories to help find the best match with an organization.

Specifically, suppose the organization text is denoted by $o_i$, and each category description is denoted by $d_j$ ($j = 1..n$). We created *n* sequence pairs $(o_i, d_1)$, $(o_i, d_2)$, ..., $(o_i, d_n)$ and asked a system to assign a number between 0 and 1 to each pair $(o_i, d_j)$ representing the strength of association between organization $o_i$ and category $d_j$. The OrgModel-2 model takes the text for an organization and $n$ label descriptions and predicts a strength value for each $(o_i, d_j)$ pair. Figure 2 depicts the full architecture of the OrgModel-2 model.

## 4. Evaluation

### 4.1. Evaluation Metrics

To evaluate the ability of our NLP models to classify organizations, we report three evaluation metrics that are commonly used in the NLP research community: *precision*, *recall* and *f1-score*. Intuitively, precision captures the accuracy when predicting a category, while recall captures coverage for recognizing instances of the category. These metrics are defined with respect to a specific category $C$ and computed as: $precision = TP/(TP + FP)$, $recall = TP/(TP + FN)$, $f1\text{-}score = 2/(precision^{-1} + recall^{-1})$, where **TP** (true positive) is the number of true instances of $C$ that were correctly identified by the model; **FP** (false positive) is the number of instances that the model predicted as $C$ but do not belong to $C$; **FN** (false negative) is the number of true instances of $C$ that were not identified by the model. The f1-score is the harmonic mean of precision and recall, which is an average over precision and recall that also reflects how balanced they are (more balanced is better).

In addition, we report micro and macro averaged scores for each evaluation metric, which provides an overall view of performance across the different categories. The micro average score aggregates the results for all instances (across all classes) and then calculates the metric. This gives more weight to the most frequent classes because they have more instances. In contrast, the macro average score calculates the metric for each class separately and then averages the results, which gives equal weight to all classes. For example, micro and macro averaged precision are defined as: $micro\ precision = \sum_{i=1}^{n} TP_i / (\sum_{i=1}^{n} TP_i + \sum_{i=1}^{n} FP_i)$, $macro\ precision = \frac{1}{n}\sum_{i=1}^{n} TP_i/(TP_i + FP_i)$, where $i$ indicates the $i$-th class, and $n$ is the total number of class labels.

### 4.2. Results and Analysis

#### 4.2.1. Environmental Issues Classification Results

For our experiments, the 1,870 organizations in the environmental issues dataset were split into 1,370 for training and 500 for testing. Table 3 shows the experimental results for each environmental issue category.

**OrgModel-1 vs. OrgModel-2** Our OrgModel-1 system achieved 76.8% micro-averaged f1-score and 69.6% macro-averaged f1-score. By adding the environmental issue category descriptions, the OrgModel-2 achieved better performance with a 80.1% micro-averaged f1 and 74.0% macro-averaged f1. Overall, we see good precision for most of the categories (87% micro-average), but recall is lower (74% micro-average).

If we take a closer look at the individual categories, we can see that OrgModel-2 achieved the best performance (>80% f1-score) for *Biodiversity*, *Governance*, *Physical*

|  | OrgModel-1 | | | OrgModel-2 | | |
| --- | --- | --- | --- | --- | --- | --- |
|  | Precision | Recall | F1-score | Precision | Recall | F1-score |
| Air & Climate | 69.2 | 32.1 | 43.9 | 75.0 | 53.6 | 62.5 |
| Biodiversity | 80.6 | 84.5 | 82.5 | 85.6 | 80.4 | 82.9 |
| Common Pool Resources | 57.1 | 40.0 | 47.1 | 70.0 | 46.7 | 56.0 |
| Disasters | 85.7 | 44.4 | 58.5 | 68.4 | 48.1 | 56.5 |
| Food Production | 66.7 | 47.3 | 55.3 | 81.2 | 60.2 | 69.1 |
| Governance | 97.5 | 88.6 | 92.9 | 100.0 | 94.3 | 97.1 |
| Institutions | 85.9 | 62.6 | 72.4 | 85.5 | 66.4 | 74.7 |
| Land & Soil | 74.4 | 75.6 | 75.0 | 77.9 | 72.5 | 75.1 |
| Physical Infrastructure | 94.0 | 83.0 | 88.1 | 95.4 | 88.8 | 92.0 |
| Protected Areas | 77.2 | 57.9 | 66.2 | 81.7 | 64.5 | 72.1 |
| Public Health | 72.0 | 40.9 | 52.2 | 58.3 | 47.7 | 52.5 |
| Sociocultural Systems | 83.3 | 56.6 | 67.4 | 85.3 | 54.7 | 66.7 |
| Technology | 100.0 | 66.7 | 80.0 | 100.0 | 71.8 | 83.6 |
| Wastes & Pollution | 83.7 | 71.1 | 76.9 | 87.4 | 76.3 | 81.5 |
| Water | 93.2 | 79.2 | 85.6 | 94.4 | 81.4 | 87.4 |
| Micro Average | 84.5 | 70.4 | 76.8 | 87.2 | 74.2 | 80.1 |
| Macro Average | 81.4 | 62.0 | 69.6 | 83.1 | 67.2 | 74.0 |

**Table 3.** Environmental issues classification: OrgModel-1 and OrgModel-2 performance for each category.

*Infrastructure*, *Technology*, *Wastes & Pollution*, and *Water*. The model struggled the most (< 60% f1 score) for *Common Pool Resources*, *Disasters*, and *Public Health*.

**Data Sparsity** We investigated whether the number of instances has an impact on the model performance. We focused on the 5 least common categories which are associated with fewer than 10 percent of all organizations in the dataset: *Public Health*, *Disasters*, *Common Pool Resources*, *Air & Climate* and *Technology*; and the 5 most common categories which are associated with more than 20 percent of all organizations: *Water*, *Physical Infrastructure*, *Wastes & Pollution*, *Biodiversity*, and *Land & Soil*. Figure 3 shows a plot of their f1-scores for both OrgModel-1 and OrgModel-2. Comparing rare categories (circles) with common categories (squares), we can see that the latter perform better with both OrgModel-1 and OrgModel-2. OrgModel-1's average f1-score of the common categories is 81.6% but the average f1-score of rare categories is only 56.3%. OrgModel-2's average f1-score of the common categories is 83.8% and the average f1-score of rare categories is 62.2%. This huge gap indicates that performance on the rare categories will likely improve if we can get more human-annotated data.

**Component-Integrated Mapping** During our dataset construction for environmental issue classification, we map component issues to integrated issues to focus on a limited set of common labels. However, this many-to-many mapping can create some problems. For example, the component issue *Land Use* is associated with the integrated issue *Food Production*. Yet not all organizations tagged with *Land Use* should be tagged with *Food Production*, e.g., organizations "Air Force Civil Engineer Center" and "Snowlands Network" don't actually produce any food. For future work, we will consider a better alignment between component and integrated issues, or refine the label set by including component issues directly.

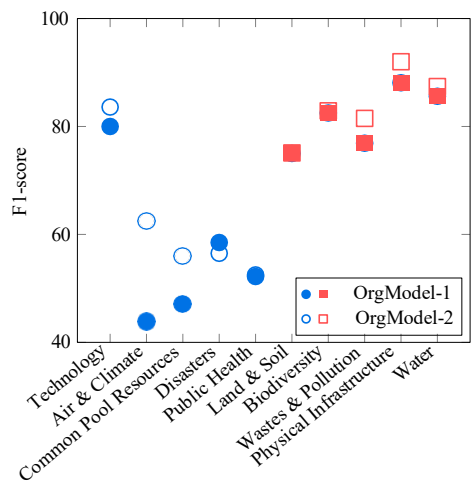

**Figure 3.** F1-score for each category. Circles ◯ represent low frequency environmental issues and squares ☐ represent high frequency issues.

*4.2.2. SIC Code Classification Results*

For the SIC code classification task, we split the 5,400 organizations into train, development, and test sets containing 2700, 900, and 1800 examples respectively.

**10-K Filing vs. Google PseudoDoc** Our first experiment explores the use of two different types of textual data for the SIC code classification task: the Pseudo-documents created from Google's retrieved text snippets, and the 10-K filing forms from the SEC database. We trained two different OrgModel-1 systems, one using textual data from the pseudo-documents and one using the 10-K filing reports. These NLP models are identical except for the textual descriptions used in the training process. Table 4 shows that the model which used Google snippets as the text source for an organization strongly outperformed the model which used the text from 10-K forms.[9] In fact, the model trained on Google snippets achieved a higher f1-score across every single category except for one (62 - Security And Commodity Brokers, Dealers, Exchanges, And Services). This category had the smallest difference in performance between the two models at 1.3%. In many cases, the model trained with Google snippets outperformed the model trained with 10-K forms by a significant margin leading to a 13% increase in macro-averaged f1-score. This approach is also more generalizable because many organizations do not have SEC filings.

**OrgModel-1 vs. OrgModel-2** We trained both OrgModel-1 and OrgModel-2 with Google PseudoDoc texts for the SIC code classification task as well. For OrgModel-2, we also need a description for the category labels, and we experimented with 3 different types of information: 1) The shortest representation (**Short**) is simply the name of the 2 digit Major Group. 2) The second representation (**Tree**) exploits the hierarchical nature of SIC codes by concatenating together the 2-digit name and the names for all the codes in its subtree. The motivation here is to include the specific subcategory information to create a richer representation of the parent category. 3) We used the plain text explanation (**Long**) of the Major Group found in the SIC manual. This is a written explanation of the criteria for an entity to be included in the specified Major Group. Table 5 shows the results. The short representation didn't contain enough information to perform well. Between the tree and long representations, we see similar performance in terms of f1-score. Comparing OrgModel-2 with tree description to OrgModel-1, OrgModel-2 achieved a 3% increase in f1-score making it our best performing model on the SIC task.

---

[9]It is worth noting that about 4% of the organizations' 10-K filings extracted are empty.

|  | 10-K Filing | | | Google PseudoDoc | | |
|---|---|---|---|---|---|---|
| Class | Recall | Precision | F1-Score | Recall | Precision | F1-Score |
| 10 | 54.8 | 73.9 | 62.9 | 80.2 | 82.6 | 81.4 |
| 13 | 67.7 | 66.6 | 67.1 | 82.8 | 84.1 | 83.4 |
| 20 | 54.2 | 83.6 | 65.8 | 77.4 | 90.1 | 83.3 |
| 27 | 67.1 | 70.4 | 68.8 | 72.5 | 73.7 | 73.1 |
| 28 | 71.4 | 60.8 | 65.6 | 71.7 | 75.6 | 73.6 |
| 34 | 38.0 | 45.0 | 41.2 | 68.5 | 52.1 | 59.2 |
| 35 | 24.2 | 11.1 | 15.2 | 46.3 | 44.4 | 45.3 |
| 36 | 28.5 | 36.6 | 32.1 | 43.0 | 51.6 | 46.9 |
| 37 | 47.1 | 57.8 | 51.9 | 68.8 | 73.6 | 71.1 |
| 38 | 63.3 | 61.2 | 62.2 | 59.4 | 75.8 | 66.6 |
| 48 | 66.6 | 58.1 | 62.1 | 73.5 | 70.9 | 72.2 |
| 49 | 68.9 | 80.9 | 74.4 | 74.6 | 84.1 | 79.1 |
| 50 | 32.2 | 15.1 | 20.6 | 73.6 | 42.4 | 53.8 |
| 51 | 28.1 | 13.4 | 18.1 | 47.4 | 41.7 | 44.4 |
| 58 | 89.2 | 75.3 | 81.6 | 86.4 | 83.1 | 84.7 |
| 59 | 48.4 | 40.7 | 44.2 | 68.1 | 59.2 | 63.3 |
| 60 | 81.5 | 84.1 | 82.8 | 90.6 | 92.0 | 91.3 |
| 61 | 70.0 | 71.0 | 70.5 | 89.3 | 85.5 | 87.4 |
| 62 | 77.2 | 85.9 | 81.3 | 75.9 | 84.5 | 80.0 |
| 63 | 55.9 | 91.0 | 69.3 | 91.0 | 91.0 | 91.0 |
| 65 | 66.1 | 66.1 | 66.1 | 69.5 | 70.5 | 70.0 |
| 67 | 49.3 | 50.7 | 50.0 | 52.1 | 52.1 | 52.1 |
| 70 | 71.4 | 69.4 | 70.4 | 85.7 | 83.3 | 84.5 |
| 73 | 28.0 | 26.6 | 27.3 | 30.0 | 35.0 | 32.3 |
| 79 | 67.3 | 42.3 | 51.9 | 75.7 | 64.1 | 69.4 |
| 80 | 73.7 | 70.3 | 72.0 | 81.8 | 84.3 | 83.0 |
| 87 | 30.5 | 41.2 | 35.1 | 54.2 | 60.3 | 57.1 |
| Macro Avg | 56.4 | 57.4 | 56.8 | 70.0 | 69.9 | 69.9 |

**Table 4.** Performance of OrgModel-1 trained on different data sources.

|  | Label Desc. | Precision | Recall | F1 |
|---|---|---|---|---|
| OrgModel-1 | - | 69.9 | 70.0 | 69.9 |
| OrgModel-2 | Short | 47.0 | 49.9 | 46.1 |
|  | Tree | 73.6 | 73.6 | 73.4 |
|  | Long | 73.6 | 73.5 | 73.1 |

**Table 5.** Macro average scores for OrgModel-1 and OrgModel-2 with different representations for the label description.

Overall, the classifier achieves reasonable performance across the different categories; however, there are a few categories (35, 50, 51, 67) where the performance is still weak, possibly due to being particularly broad and difficult to distinguish from other categories. Yet, our model achieves greater than 70% f1-score on the majority of categories, illustrating the promise of using NLP models to automatically classify organizations.

**5. Conclusion and Future Plans**

Our study has demonstrated the potential of using Natural Language Processing (NLP) techniques for the automatic acquisition of structured food system knowledge from unstructured sources. This includes the classification of entities involved in food activities, and their linkage to social, environmental, and health-related issues. Using text snippets retrieved from Google's search engine as a descriptive basis, our NLP models were able to classify organizations according to environmental issue categories and Standard Industrial Classification (SIC) codes. Specifically, we build our models based on transformer-based language models. Our experimental results show that using textual data from Google Search achieved a better performance than using 10-K filings from organizations' annual reports, which are provided by the Securities and Exchange Commission (SEC) database. We also show that by incorporating the description text of the environmental issue categories or SIC codes, our model can achieve better performance for both classification tasks. The promising results underline the applicability of this approach to improve food ontologies as well as other intelligent food systems knowledge resources with little to no human supervision. Our work illustrates how NLP models hold the potential for a wide range of automatic categorization of food system actors and their activities into existing food, environment, and health system ontologies. These ontologies instantiated by NLP models represent burgeoning yet powerful instruments for quick and efficient population of large food systems knowledge graphs with consistent knowledge representation.

Our research opens the door to providing a critical component to the design and creation of reusable cyberinfrastructure components capable of addressing major social-environmental issues like scaling climate actions in food systems. Other components will include the development of additional technological, knowledge, and relational infrastructures that will build on recent advances in data, modeling, and management. Linking these components together affords the opportunity to make leading edge insights, tools, and practical guidance available to action partners across the food system.

**Acknowledgments**

We want to thank the anonymous reviewers for their valuable comments. This research was supported in part by the ICICLE project through NSF award OAC 2112606 and the Canadian Institutes of Health Research (CIHR) FRN 177412.